\documentclass[pdflatex,sn-mathphys-num]{sn-jnl}

\usepackage{graphicx}%
\usepackage{multirow}%
\usepackage{amsmath,amssymb,amsfonts}%
\usepackage{amsthm}%
\usepackage{mathrsfs}%
\usepackage[title]{appendix}%
\usepackage{xcolor}%
\usepackage{textcomp}%
\usepackage{manyfoot}%
\usepackage{booktabs}%
\usepackage{algorithm}%
\usepackage{algorithmicx}%
\usepackage{algpseudocode}%
\usepackage{listings}%
\usepackage{tabularx}  
\usepackage{enumitem}

\usepackage{makecell}
\usepackage{tabularx}
\usepackage{booktabs}
\usepackage{array}

\usepackage{booktabs}
\usepackage{array}

\theoremstyle{thmstyleone}%
\theoremstyle{thmstyletwo}%

\theoremstyle{thmstylethree}%

\begin{document}

\title[DyABD]{DyABD: The Abdominal Muscle Segmentation in Dynamic MRI Benchmark}


\author*[1,2]{\fnm{Niamh} \sur{Belton}}\email{niamhbelton@gmail.com}
\author[3,4]{\fnm{Victoria} \sur{Joppin}}
\author[5,6]{\fnm{Aonghus} \sur{Lawlor}}
\author[3]{\fnm{Catherine} \sur{Masson}}
\author[3,7]{\fnm{Thierry} \sur{Bege}}
\author[4]{\fnm{David} \sur{Bendahan}}
\author[1,2,6]{\fnm{Kathleen M.} \sur{Curran}}

\affil[1]{\orgdiv{School of Medicine}, \orgname{University College Dublin}, \orgaddress{\city{Dublin}, \country{Ireland}}}
\affil[2]{\orgdiv{Science Foundation Ireland Centre for Research Training in Machine Learning},  \orgaddress{\city{Dublin}, \country{Ireland}}}
\affil[3]{\orgdiv{Aix Marseille Univ, Univ Gustave Eiffel, LBA, Marseille France}}
\affil[4]{\orgdiv{Aix Marseille Université, CNRS, CRMBM UMR 7339, Marseille France}}
\affil[5]{\orgdiv{School of Computer Science}, \orgname{University College Dublin}, \orgaddress{\city{Dublin}, \country{Ireland}}}
\affil[6]{\orgdiv{Insight Centre for Data Analytics, University College Dublin, Dublin, Ireland}}
\affil[7]{\orgdiv{Aix Marseille Université, Département de Chirurgie, Hôpital Nord, APHM, Marseille, France}}


\abstract{
This work introduces \textbf{DyABD}, a novel and complex benchmark dataset of dynamic abdominal MRIs from patients with abdominal hernias and associated high quality abdominal muscle annotations. DyABD is the first-of-its-kind in four key ways; (1) it proposes the first abdominal muscle segmentation task, (2) the dynamic MRIs are acquired whilst the patients perform various exercises, introducing extreme anatomical variability, making it one of the most challenging segmentation datasets to date, (3) it includes both pre and post corrective MRIs and (4) DyABD promotes clinical research into the high recurrence rates of abdominal hernias. Beyond dataset introduction, this work provides a comprehensive evaluation of the generalisation capabilities of existing segmentation models across Supervised, Few Shot and Zero Shot paradigms on the unseen DyABD dataset. This work reveals that there is still room for substantial improvement in the field of medical image segmentation, with the majority of techniques achieving a Dice Coefficient of 0.82. This work therefore sheds light on the true progress of the field and redefines the benchmark for progress in medical image segmentation. }




\maketitle

\section{Introduction}

\begin{figure}[H]
    \centerline{\includegraphics[width=\textwidth]{examples2.pdf}}

    \caption[Examples of Dynamic MRIs from the DyABD Dataset]{(a) Examples of three patients performing each of the exercises, breathing, coughing and Valsalva maneuver at the beginning of the cycle, mid cycle and the end of the cycle at the preoperative stage. (b) The same patients at the postoperative stage. All examples are shown with annotations of the abdominal muscles, right Lateral Muscle (lime green), right Rectus Abdominus (dark green), left Rectus Abdominus (grey) and left Lateral Muscle (purple). Red arrow indicates the position of the abdominal hernia.}
    \label{fig:fss_challenge_examples}
    \vspace{0.06cm}
\end{figure}

This work introduces a complex and challenging medical imaging benchmark dataset, \textbf{DyABD}\footnote{Dataset available at \url{https://entrepot.recherche.data.gouv.fr/dataset.xhtml?persistentId=doi:10.57745/KTM2OA}. Code and visuals available at \url{https://github.com/niamhbelton/DyABD-Segmentation}.}. DyABD consists of 311 dynamic abdominal MRIs from patients with abdominal hernias pre-corrective and post-corrective surgery, acquired whilst each patient performed three distinct exercises; breathing, coughing and the Valsalva maneuver. Following a rigorous two-year acquisition and annotation process, the dataset is complete with associated high-quality annotations of four abdominal muscle groups. DyABD is both novel and distinctive for the following four reasons;

\begin{enumerate}
    \item \textbf{First-of-its-kind task:} Whilst abdominal organ segmentation in static MRIs has previously been studied \citep{fewshotsuper,fss_recurrent,adnet,dual_contrast,cycle_resem,oneshotfss,fss_transformer,fss_mult}, this benchmark dataset is the first to address the unexplored task of abdominal muscle segmentation in dynamic MRIs for patients with abdominal hernias.
\item \textbf{Unique and challenging segmentation task:} The dynamic nature of the dataset introduces substantial anatomical deformation and muscle displacement, making it one of the most complex and challenging publicly documented segmentation tasks to date.
\item \textbf{Clinical significance:} This dataset directly supports research addressing a critical clinical issue, high recurrence rates of abdominal hernias following surgical repair \citep{recurr}. Understanding the underlying causes of recurrence requires detailed analysis of the mechanical functionality of the abdominal wall. By enabling the development of automatic abdominal muscle segmentation models, DyABD makes it feasible to perform such analyses at scale, paving the way to improved understanding of abdominal wall biomechanics and thus, improved clinical decision making.
\item \textbf{DyABD's broader impact:} A key novelty of DyABD is its inclusion of MRIs acquired both pre and post-corrective surgery. DyABD therefore is not only valuable for developing accurate segmentation models, but also for other clinically important tasks such as clinical outcome prediction and personalised treatment planning. 
\end{enumerate}

In addition to the introduction of the DyABD dataset, this work demonstrates that widely used medical image segmentation models fall short when tested on an unseen and challenging dataset. A comprehensive evaluation of these models’ generalisation capabilities, performed as part of this work, reveals that current approaches have limited capabilities outside of standard benchmark settings. This analysis therefore sheds light on the true progress of the field. Given the small-medium size of the dataset, this work in particular focuses on the setting where there is a scarcity of training data available, a common scenario in medical image analysis. This work assesses segmentation models across three paradigms;

\begin{enumerate}

    \item Supervised approaches \citep{umamba,nnunet} which are designed to train on large annotated datasets.
    \item Few Shot Segmentation (FSS) approaches \citep{adnet,universeg} which are designed to train on large annotated datasets but can make predictions on unseen classes at test time given \lq{}few\rq{} examples.
    \item  Zero Shot approaches \citep{sam,medsam} which require no further fine tuning or training but require a prompt in the form of a bounding box, key points, text or other prompting methods. 
\end{enumerate}

 This analysis reveals that off-the-shelf Zero Shot foundation models SAM \citep{sam} and SAM 2 \citep{sam2} can outperform specialised models like MedSAM \citep{medsam}, which have been fine-tuned on medical imaging data. This finding highlights the ongoing need for research into the development of universal, generalisable Zero-Shot models for medical image segmentation. Furthermore, this work reveals that despite the progress of Few Shot and Zero Shot techniques, a fully supervised 3D model, nnU-Net, continues to perform on par with these approaches in the limited data regime.

Section \ref{sec:dyabd} describes the DyABD dataset, while sections \ref{sec:exp} to \ref{sec:res} outline the comprehensive evaluation of the performance of current segmentation models.

\section{Motivation}

\subsection{Clinical Relevance}
The dataset was acquired for the pioneering analysis of the biomechanical functionality of the abdominal wall in patients with abdominal hernias from dynamic MRI. Abdominal hernias are characterised by the protrusion of viscera between a rupture in the abdominal wall muscles \citep{hernia}, causing debilitating discomfort to the patient and ongoing clinical challenges. The recurrence rate of abdominal hernia post corrective surgery varies from 30\% to 80\% \citep{recurr}, resulting in increased healthcare costs and longer patient waiting lists. Improving treatment outcomes requires a deeper understanding of abdominal wall mechanics and physiology. Measuring key metrics such as the radial displacement \citep{dynamic_mri,vic} of the abdominal muscles, right Lateral Muscle (rLA), right Rectus Abdominus (rRA), left Rectus Abdominus (lRA) and left Lateral Muscle (lLM) as the patient performs various exercises can provide valuable insights into muscle functionality. However, obtaining such statistics requires the precise segmentation of each of the abdominal muscles. This is often performed manually which is a tedious and time consuming process. Dynamic MRI was the modality chosen for this study as recent works have shown this modality to be effective in quantifying abdominal wall motion and deformation during various exercises such as breathing and muscular contraction \citep{dynamic_mri}. To the best of the author's knowledge, this is the first work to perform the automatic segmentation of abdominal muscles for patients with abdominal hernias from dynamic MRI.

\subsection{Comparison to Few Shot Segmentation Benchmark Datasets}
There exists open-source medical imaging benchmark datasets specifically for the purposes of training and testing FSS models. Table \ref{tab:rel_fss_datasets} presents the most common of these datasets and their characteristics. The table highlights each dataset, the modality, region of the body, number of volumes that each work used in experiments and the associated task. It also lists the works that developed an FSS model by training and testing on these datasets. It can be observed from the table that the vast majority of FSS methods are trained and tested on the task of organ segmentation in abdominal MRI and CT, as well as cardiac segmentation in MRI. There is a notable gap in assessing the generalisability of these models to other tasks within the field such as abdominal muscle segmentation.

\begin{table}[htbp]
\centering
\caption{Summary of Few Shot Segmentation benchmark datasets and the DyABD dataset with their associated characteristics.}
\vspace{0.2cm}
\label{tab:rel_fss_datasets}

\begin{tabularx}{\textwidth}{
    >{\centering\arraybackslash}p{2.0cm} 
    >{\centering\arraybackslash}p{1.4cm} 
    >{\centering\arraybackslash}p{1.6cm} 
    >{\centering\arraybackslash}p{1.4cm} 
    >{\centering\arraybackslash}p{2.6cm} 
    >{\raggedright\arraybackslash}X
}
\toprule
\toprule
\makecell{\textbf{Dataset}} & \textbf{Modality} & \textbf{Anatomy} & \textbf{\#Volumes} & \textbf{Task} & \makecell{\textbf{Related} \\ \textbf{Works}} \\
\midrule

\makecell[l]{ABD-30 \citep{abd30}} & CT & Abdomen & 30 & \makecell{Organ \\ Segmentation} & \scriptsize \citep{fewshotsuper,fss_recurrent,ouyang2,dual_contrast,cycle_resem,fss_transformer,fss_mult} \\

\makecell[l]{MMWHS \citep{mmwhs}} & CT\&MRI & Cardiac & 20 & \makecell{Cardiac \\ Segmentation} & \scriptsize \citep{local_cont} \\

\makecell[l]{Card-MRI \citep{card_mri}} & MRI & Cardiac & 35 & \makecell{Cardiac \\ Segmentation} & \scriptsize \citep{fewshotsuper,ouyang2,adnet,fss_transformer,fss_mult} \\

\makecell[l]{ACDC \citep{acdc}} & MRI & Cardiac & 150 & \makecell{Cardiac \\ Segmentation} & \scriptsize \citep{local_cont} \\

\makecell[l]{KiTS19 \citep{kits}} & CT & Abdomen & 300 & \makecell{Organ \\ Segmentation} & \scriptsize \citep{interactive} \\

\makecell[l]{ABD-MR \citep{chaos}} & MRI & Abdomen & 20 & \makecell{Organ \\ Segmentation} & \scriptsize \citep{fewshotsuper,fss_recurrent,adnet,dual_contrast,cycle_resem,oneshotfss,fss_transformer,fss_mult} \\

\makecell[l]{Medical \\ Segmentation \\ Decathlon \\ (Prostate) \citep{med_decath}} & MRI & Prostate & 48 & \makecell{Prostate \\ Structure \\ Segmentation} & \scriptsize \citep{local_cont} \\

\makecell[l]{LiTS \citep{lits}} & CT & Abdomen & 130 & \makecell{Liver Tumor \\ Segmentation} & \scriptsize \citep{interactive} \\
\hline
\makecell[l]{\textbf{DyABD}} & \textbf{Dynamic MRI} & \textbf{Abdomen} & \textbf{311} &  \makecell{Abdominal Muscle \\ Segmentation} & -- \\

\bottomrule
\end{tabularx}
\end{table}

\section{DyABD Dataset}\label{sec:dyabd}

\subsection{Acquisition}
\begin{figure*}[!t]
\centerline{\includegraphics[width=\textwidth]{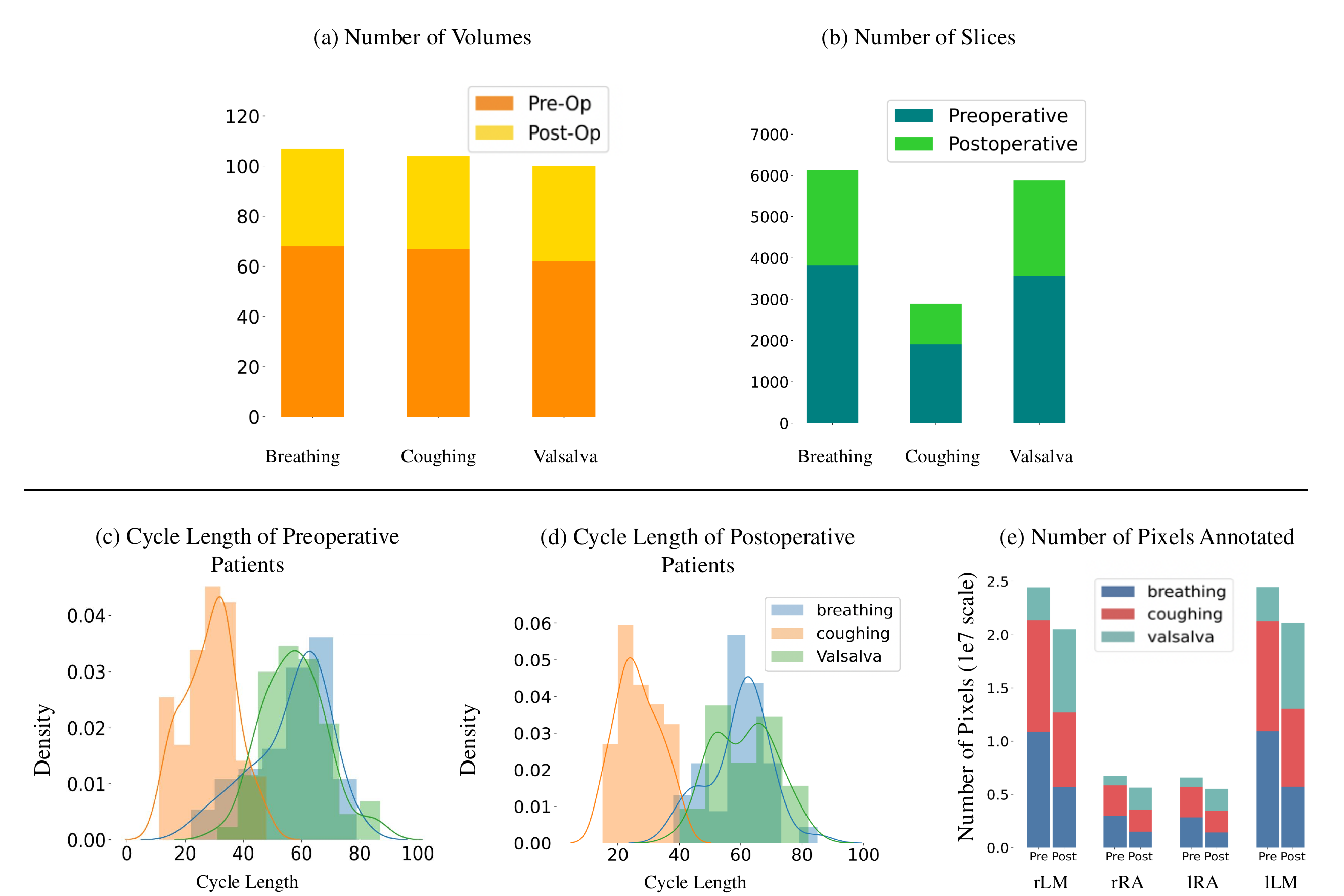}}
\caption[Further DyABD Dataset Statistics]{(a) The number of abdominal dynamic MRI volumes for each exercise broken down by pre and postoperative. (b) The number of slices for each exercise broken down by pre and postoperative. (c) A density plot of the cycle lengths (i.e. number of slices) for each exercise for preoperative patients. (d) A density plot of the cycle lengths (i.e. number of slices) for each exercise for postoperative patients. (e) The number of pixels annotated for each muscle in the dataset broken down by pre and postoperative and the exercise. }
\vspace{0.06cm}
\label{fig:fss_dataset_stats}

\end{figure*}

The dataset was acquired by collaborators of this work. 
This study was approved by the French ethics committee (IDRCB: 2021-A02119-32) and was conducted at Assistance Publique des Hôpitaux de Marseille (APHM), France according to national legislation related to interventional research and the Declaration of Helsinki. The dataset, DyABD consists of dynamic MRIs from 17 patients with abdominal hernias. Each patient performed three audio-guided exercises, breathing, coughing and Valsalva maneuver. The Valsalva maneuver can be described as attempting to push air out with both the nose and mouth closed. During acquisition, each patient performs each exercise on average four times. For each 3D volume, the 2D slice where the size of the rupture in the tissue is largest was selected. The axial slice at this position over time was collated for this dataset, resulting in a 3D volume consisting of 2D axial slices over time. A repeat dynamic MRI was acquired post-corrective surgery for ten of the aforementioned patients. The acquisition MRI parameters are outlined in table \ref{tab:fss_mri}. This resulted in MRIs with a resolution of 416 $\times$ 416. The complete dataset therefore consists of 311 volumes. Table \ref{tab:fss_stats} presents the statistics of the dataset. Figure \ref{fig:fss_dataset_stats} (a) illustrates the total number of volumes and figure \ref{fig:fss_dataset_stats} (b) shows the number slices broken down by the exercise and the operative stage. Figure \ref{fig:fss_dataset_stats} (c) and (d) show the distribution of the cycle lengths (i.e. number of slices in each volume) for preoperative and postoperative patients respectively. Due to its abrupt motion, the length of a coughing cycle is substantially shorter than breathing or Valsalva maneuver. 

\begin{table*}
  \caption[DyABD Acquisition Parameters.]{DyABD acquisition parameters.}
  \resizebox{\textwidth}{!}{

  \label{tab:fss_mri}
  \setlength{\tabcolsep}{3pt}

  \centering
  \begin{tabular}{l|cc}
    \hline
    \hline
  Exercise &  Average &  Range [min/max]            \\
  \hline

    Duration between surgery and post-operative MRI (days) & 137 $\pm{}84$ & [76 /362] \\ 
    Field of View (FoV) (mm) & 409 $\pm{} 31$ & [360 / 470] \\
    Temporal resolution (ms) & 160 $\pm{} 4$ & [154 / 166] \\ 
    Spatial resolution (mm) & 0.983 $\pm{} 0.075$ & [0.865 / 1.130] \\ 
        
    \hline
  \end{tabular}}
\end{table*}

\begin{table*}
  \caption[DyABD Dataset Statistics.]{DyABD dataset statistics.}
  \resizebox{\textwidth}{!}{

  \label{tab:fss_stats}
  \setlength{\tabcolsep}{3pt}

  \centering
  \begin{tabular}{l|cccc|cccc}
    \hline
  Exercise &  \multicolumn{4}{|c|}{Pre-Op}  &  \multicolumn{4}{c}{Post-Op}                  \\
         & \#Volumes  & \#Slices  & \#Patients & \#Annotations & \#Volumes & \#Slices    & \#Patients & \#Annotations\\
         &&&&(pixels) &&&&(pixels)\\
    \hline
    Breathing     & 68  & 3,815 &17& 27,650,739& 39 &2,317 &10 & 18,390,091\\
    Coughing & 67 & 1,909 &17& 14,324,548 &37 & 979 &10 & 8,113,246\\
    Valsalva   & 62 & 3,565 &17& 26,446,718 &38 &2,323 &10 & 20,029,488\\
    \hline
  \end{tabular}}
\end{table*}

\subsection{Annotation}
\begin{figure*}[!t]
    \centering
    \includegraphics[width=\textwidth]{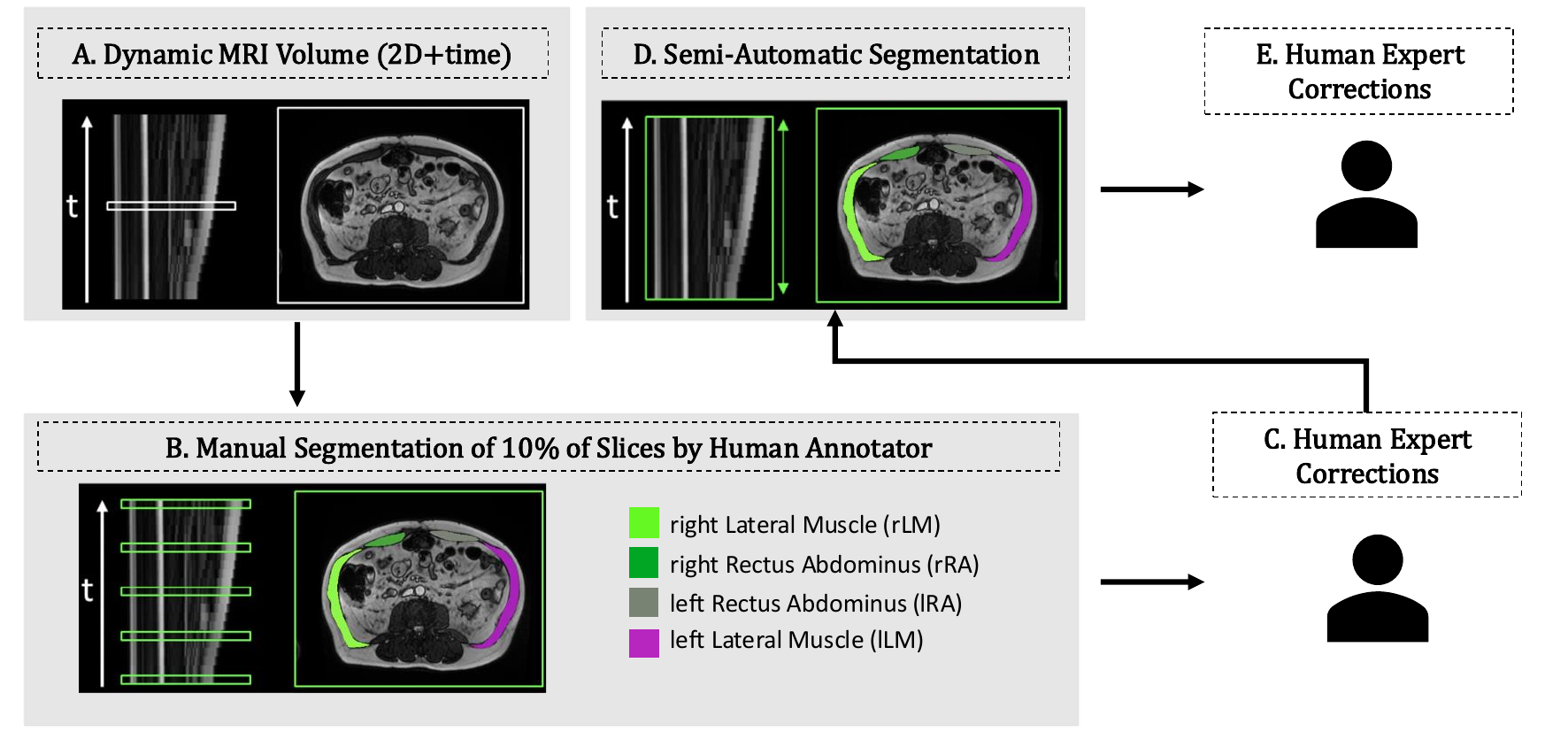}
    \caption[Annotation Pipeline]{A. The figure shows a dynamic MRI volume of 2D axial slices over time. The image indicates where the shown axial slice is located in the time dimension. B. A human annotator, the author of this work, annotated approximately 10\% of the slices. C. The annotations are reviewed and corrected by a human expert. D. The semi-automatic tool is used to propagate the annotations to the remaining slices. E. These are reviewed again and corrected by a human expert. }
    \label{fig:fss_ann_pipe}
    \vspace{0.06cm}
\end{figure*}

Figure \ref{fig:fss_ann_pipe} presents the annotation pipeline. The annotation process was conducted by two authors of this work, each referred to as the \lq{}human annotator\rq{} and the \lq{}human expert\rq{}. Following the acquisition of the dynamic MRIs, the human annotator annotated the following abdominal muscles, the right Lateral Muscle (rLA), right Rectus Abdominus (rRA), left Rectus Abdominus (lRA) and left Lateral Muscle (lLM). The location of each of the muscles can be seen on figure \ref{fig:fss_ann_pipe} (B). The protocol for the manual segmentation by the human annotator was as follows. The first slice in each dynamic MRI volume was manually segmented. This slice is denoted as $s$ and it is at rest position. The slice at the maximum of the exercise cycle, where the displacement of the RA muscles is at a maximum, was also segmented, along with the slice at the end of the cycle. These slices are denoted as $m$ and $e$ respectively. Additionally, the slice mid way to the maximum of the cycle, $\frac{m-s}{2}$ and mid way from the maximum to the end, $m + \frac{e-m}{2}$ was manually segmented. This corresponds to approximately 10\% of slices. It takes on average 20 minutes for the human annotator to complete the annotation of 10\% of slices of one dynamic MRI volume. The human expert, a collaborator who is a mechanical engineer with a PhD in the study of the biomechanical behaviour of the herniated abdominal wall using dynamic MRI, reviewed each of the annotated slices and made corrections if necessary. Cohen's Kappa Coefficient between the masks completed by the original human annotator and the masks post corrections by the human expert was 0.984, showing that only minor corrections were required. A semi-automatic propagation tool \citep{propagation}, developed by the collaborators of this study, was then used to annotate the remaining slices in each volume. This tool works by using non-linear registration approaches. Finally, the human expert verified the final annotations as output from the propagation tool. Figure \ref{fig:fss_dataset_stats} (e) presents the number of pixels annotated for each muscle, broken down by exercise. 

To calculate the intra-rater agreement, the human expert repeated the annotation on a sample of six MRIs. They manually segmented 10\% of slices, used the semi-automatic tool to propagate the annotations and made corrections if necessary. The intra-rater agreement between these annotations and the original annotations had a Cohen's Kappa Coefficient of 0.899, showing high agreement.

\subsection{Characteristics of the Dataset}
\begin{figure}
    \centering
    \includegraphics[width=\columnwidth]{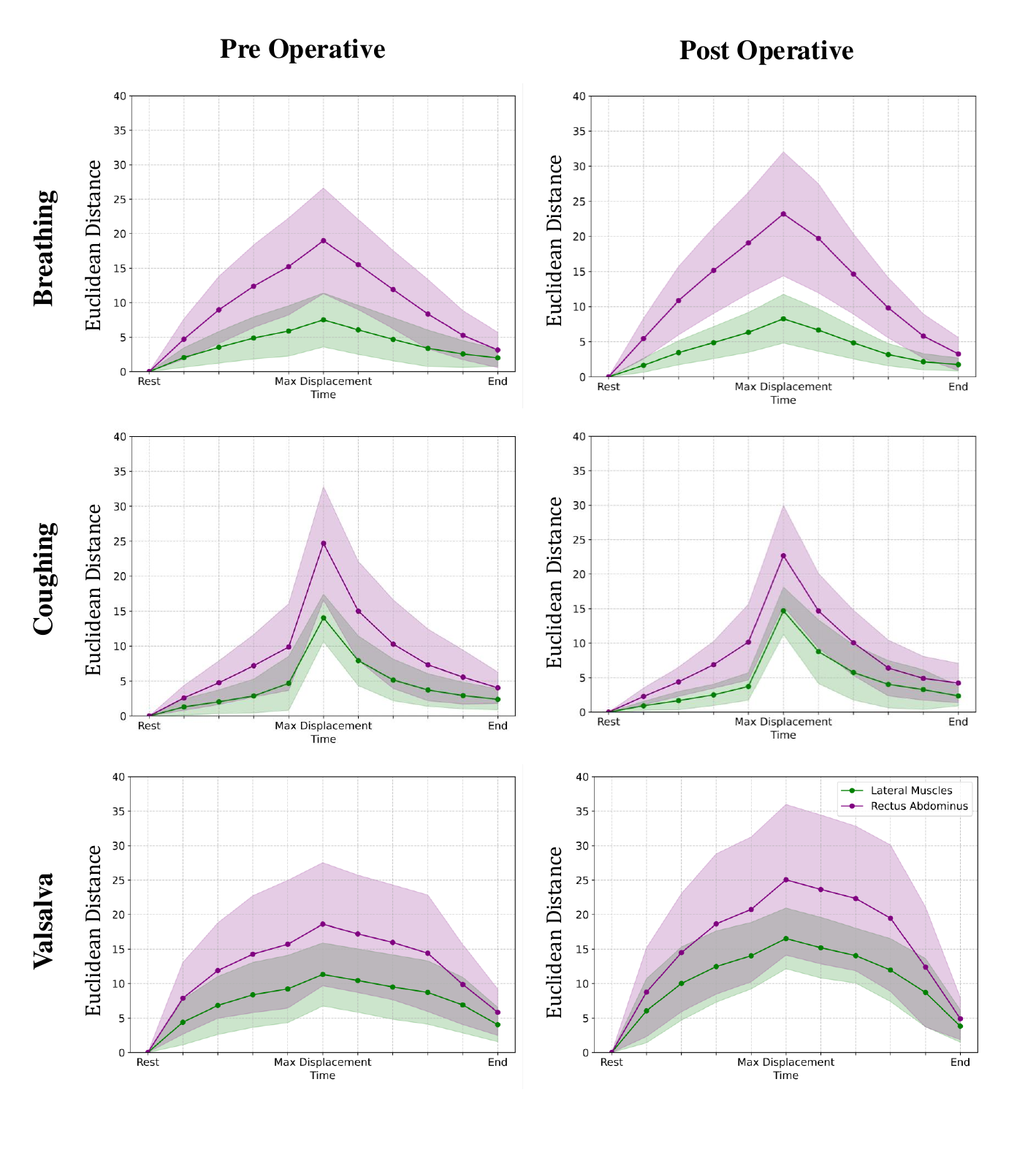}
    \caption[Level of Motion of Abdominal Muscles During Exercises.]{The Euclidean Distance (ED) of the Lateral Muscles (LM) (green) and Rectus Abdominus (RA) (purple) muscles for preoperative vs. postoperative (columns) and exercise (rows) over time.}
    \label{fig:fss_radial_dist2}
\end{figure}

Figure \ref{fig:fss_challenge_examples} shows examples of three patients at the beginning, mid and end of a cycle for each of the three exercises pre and postoperative. The annotations of each muscle are superimposed on the images. The red arrows on the figure point to where the abdominal hernia appears between the rRA and lRA muscles in the preoperative stage. High inter-patient variability can be observed from the figure, meaning MRIs of patients appear substantially different from each other due to various factors such as difference in body fat percentage, age, gender, abdominal muscle development, and hernia location. The differences in the appearance of the MRIs can be observed by comparing patients A-C row wise in figure \ref{fig:fss_challenge_examples}. Additionally, there can be high intra-patient variability within volumes due to the high range of motion in a cycle, particularly in the case of the RA muscles. To further highlight the range of motion of the abdominal muscles, figure \ref{fig:fss_radial_dist2} shows the Euclidean Distance (ED) between the barycentre of the muscle at rest and the barycentre of the muscle at all other time points. The ED of the rRA and lRA muscles were averaged, as well as the rLM and lLM muscles for visualisation. It can be observed that the RA muscles have substantially more movement than the LM muscles, indicating that they will be more challenging to segment.

\section{Experiments}\label{sec:exp}
\subsection{Overview}
The following experiments evaluate the generalisation capabilities of existing segmentation models. Specifically, the experiments evaluate three distinct approaches to the task of abdominal muscle segmentation in dynamic MRI using DL based SOTA segmentation models; 

\begin{itemize}
    \item \textbf{Semi-automatic with annotations.} This approach requires that a portion of slices from each of the dynamic MRI volumes are annotated at test time. FSS techniques are studied for this setting.
    \item \textbf{Semi-automatic with prompts.} This approach requires a prompt in the form of a bounding box for all slices of each dynamic MRI volume at test time. Promptable Zero Shot techniques are investigated for this setting.
    \item \textbf{Fully automatic.} This technique requires no level of supervision at test time. This work investigates the feasibility of supervised approaches for this setting. 
\end{itemize}

\begin{figure*}[!t]
\centerline{\includegraphics[width=\textwidth]{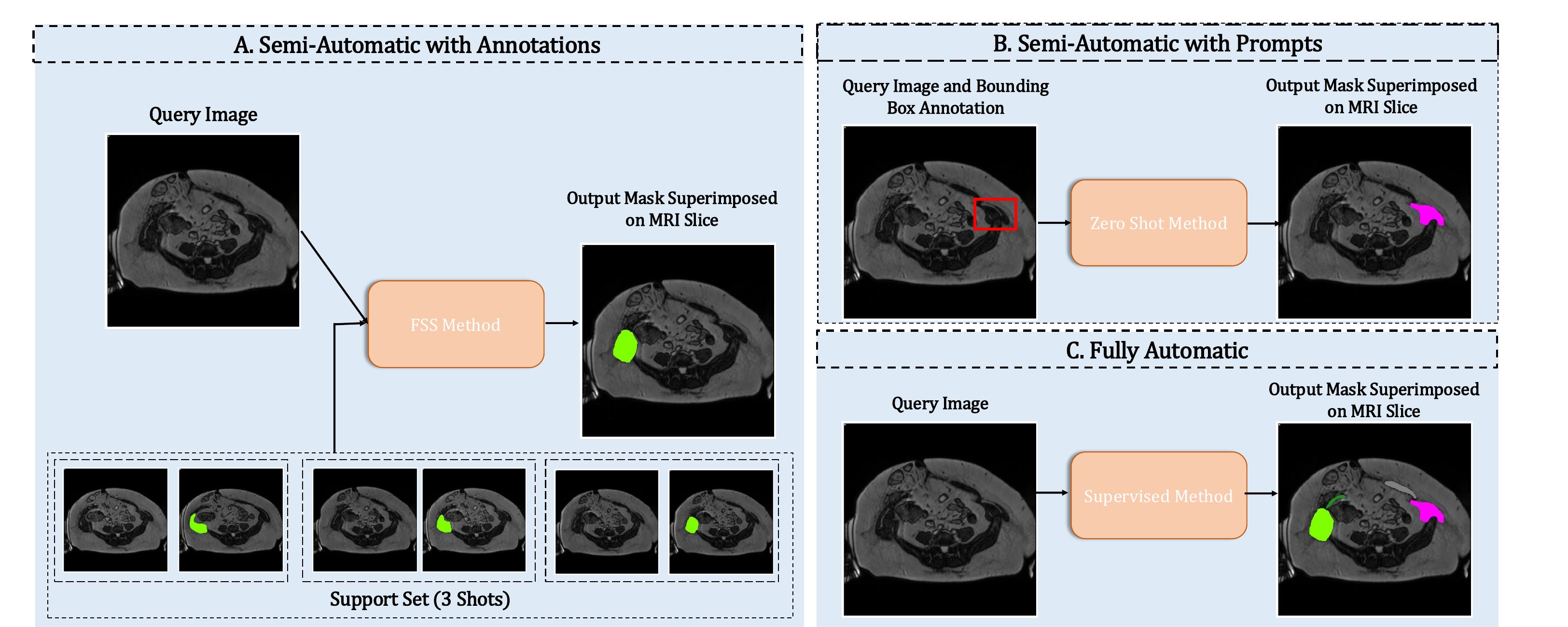}}
\caption[Diagram of Semi-Automatic and Fully Automatic Approaches at Inference Time.]{Overview of segmentation approaches: (A) Semi-automatic with annotations, requiring a portion of slice annotations at test time, known as the \lq{}Support Set\rq{} (3 \lq{}shots\rq{} (image-mask pairs) are visualised); (B) Semi-automatic with prompts, requiring bounding box prompts; and (C) Fully automatic, requiring no supervision at test time. Semi-automatic methods with annotations and semi-automatic methods with prompts perform the segmentation on a per-muscle basis. Therefore, four passes through the model for each of the abdominal muscles is required to obtain the final segmentation mask.}
\vspace{0.06cm}
\label{fig:fss_approaches}

\end{figure*}

The set-up of each approach at inference time is illustrated in figure \ref{fig:fss_approaches}. This is further explained and outlined in sections \ref{sec:fss_fss} to \ref{sec:fss_fully_auto}.

\subsection{Data Organisation}
Due to the small nature of the dataset, each model that required training was implemented with four-fold cross validation. The details of each fold are presented in table \ref{tab:fss_quantity}. Each training fold had an average of 233 volumes and each test fold had an average of 78 MRI volumes. Folds were created by stratified sampling of pre/postoperative volumes to ensure that firstly, MRIs from the same patient were within the same training or test fold and secondly, that each training and test fold had examples of both preoperative and postoperative stages. 

\begin{table}
  \caption[Number of Volumes in each Training and Test Fold.]{Number of volumes in each training and test fold. }
  \label{tab:fss_quantity}

  \centering
  \begin{tabular}{l|cc|cc|cc|cc}
    \hline
\hline
   Quantity &  \multicolumn{2}{|c|}{Fold 1}  &  \multicolumn{2}{c|}{Fold 2}  
  &  \multicolumn{2}{|c|}{Fold 3}  &  \multicolumn{2}{c}{Fold4}   
      \\
      & Train & Test & Train & Test& Train & Test& Train & Test \\
      \hline

     Patients & 12 & 5 & 13 & 4 & 13 & 4 & 13 & 4\\
    Preoperative Volumes & 138 & 59 & 149 & 48 & 152 & 45 & 152 & 45 \\ 
    Postoperative Volumes & 91 & 23 & 90 & 24 & 81 & 33 & 80 & 34\\ 
    Total Volumes & 229 & 82 &239 & 72 & 233 & 78 & 232 &79
      
  \end{tabular}
  
  \end{table}


\subsection{Semi-Automatic with Annotations}\label{sec:fss_fss}
Semi-automatic approaches substantially reduce manual annotation costs by requiring only a portion of the dynamic MRI volume to be annotated at test time. This approach leverages the information from the annotated portion to annotate the remaining slices, as can be seen from figure \ref{fig:fss_approaches} (A). FSS techniques segment semantic classes at test time in a Query Set when given a Support Set which consists of \lq{}few\rq{} examples of images with their associated segmentation mask of the semantic class. Generally FSS models aim to segment semantic classes that were previously unseen at test time. However, the experimental set-up of this work differs as the model was trained on all available semantic classes at training time.

\subsubsection{Few Shot Segmentation Preliminary}
Before outlining the FSS techniques that were investigated in this work, this section presents the FSS experimental set-up. FSS models are trained in an episodic manner, different from how typical supervised image classification or segmentation models are trained.

Consider a training fold
\[
D_{train}=\{(X_{train}^{h \times w \times slices}, \; Y_{train}^{h \times w \times slices \times c})\},
\]
which consists of dynamic MRI volumes 
$X_{train} \subseteq \mathbb{R}^{h\times w \times slices}$ 
and their associated binary masks 
$Y_{train} \subseteq \mathbb{R}^{h\times w \times slices \times c}$. 
Here, $h$ and $w$ denote the height and width of a slice, $slices$ is the number of slices in a volume, and $c$ is the number of semantic classes. The semantic classes in this case are each of the four abdominal muscles.

At training time, in episode $i$, a class $c_i \in \{1,2,3,4\}$ is randomly sampled from $c$. This is the class that the model will aim to segment in episode $i$. Each episode has a Support Set and a Query Set. The model will use the information from the labelled Support Set to segment the class in the Query Set. The Support Set of episode $i$, denoted as $S_i$, consists of $N$ slices from a single dynamic MRI volume, denoted as $x_i \subseteq \mathbb{R}^{h\times w \times N}$ and associated binary mask for class $c_i$, $y_{i,c} \subseteq \mathbb{R}^{h\times w \times N}$ so that,
\[
S_i = (x_i, y_{i,c}).
\]
The number of slices, $N$ in the Support Set is referred to as the number of \lq{}shots\rq{}. To ensure the Support Set captures muscle deformations at various stages, slices are selected at time zero and at evenly spaced intervals of \(\frac{slices}{N}\) throughout the dynamic MRI volume.

The Query Set of episode $i$, denoted as $Q_i$ consists of the remaining slices of the dynamic MRI volume, $x_i \subseteq \mathbb{R}^{h\times w \times slices-N}$. At test time, given a Support Set, $S_{test}$, a Query image $x_{test}\subseteq \mathbb{R}^{h\times w \times slices-N}$ is segmented for all classes, i.e.,
\[
\forall c \in \{1,2,3,4\}.
\]

\subsubsection{Few Shot Segmentation Techniques}

\textbf{UniverSeg.} \hspace{0.06cm} UniverSeg \citep{universeg} was investigated as it is the latest SOTA, outperforming existing FSS models such as PANet \citep{panet} and ALPNet \citep{fewshotsuper}. UniverSeg has an encoder-decoder architecture, similar to common segmentation architecture U-Net \citep{unet}. 
UniverSeg was trained by inputting the Query image into the network and inputting the Support Set which consists of both images and and associated masks into the channel dimensions with the Query image. UniverSeg employs multi-scale Cross-Blocks which interacts the representations of the Query image and Support Set at each layer. The final output is the segmentation mask of the Query image. UniverSeg requires no further training or fine tuning to other datasets due to its generalisation capabilities, which it achieved by pretraining on 53 medical imaging datasets which consist of 22,000 scans from 16 imaging modalities. The training data includes the abdominal MRI dataset for organ segmentation \citep{chaos}, the most similar open-source dataset to DyABD. For implementation on the DyABD dataset, slices were downsized from $416 \times 416$ to $128 \times 128$ and they were normalised to have pixel values between zero and one, as per UniverSeg implementation instructions. 

\textbf{ADNet.} \hspace{0.06cm} ADNet \citep{adnet} was investigated as it is one of the latest SOTA FSS models. Unlike UniverSeg, ADNet requires training on the specific dataset in question. Prior to training, slices were downsized from $416 \times 416$ to $256 \times 256$. During training, the images in the Support and Query Set were input into a ResNet-101 \citep{resnet} pretrained on the MS-COCO dataset \citep{coco}. The output representations were then used to compute a prototype of a semantic class  by using the corresponding segmentation masks to perform masked average pooling on the resultant feature map of the Support Set. An anomaly score was then assigned to each pixel in a Query image based on the negative cosine similarity to the prototype. The binary segmentation was obtained by thresholding the anomaly scores with a learned threshold parameter and upsizing the feature map to the original image size, $256 \times 256$. The original implementation of ADNet applies various transforms to either the Support Set or Query Set such as rotating, shifting, shearing or scaling. Due to the symmetry of the abdominal muscles in the MRIs, the rotating transform was removed. Additionally, while the original implementation relies on pseudo labels, the proposed implementation utilises manual labels. ADNet was trained with a learning rate of $1e-4$ and a batch size of one for 25 epochs.

\subsection{Semi-Automatic with Prompts}
Semi-automatic methods that require only a coarse bounding box annotation at test time substantially lower annotation costs, as can be seen from figure \ref{fig:fss_approaches}. 
This approach contrasts with the labour-intensive annotation process of precisely delineating the region of interest. There exists several foundation Zero Shot models that can perform segmentation when given a prompt in the form of a bounding box or key points. Similar to UniverSeg, these methods require no further retraining or fine tuning. However, they can be further fine tuned to a specific dataset or task. SAM \citep{sam}, SAM2 \citep{sam2} and MedSAM \citep{medsam} were investigated for this analysis as SAM is the SOTA Zero Shot segmentation model and MedSAM is a specialised version of SAM for medical imaging. Despite SAM being trained for natural images, \cite{sam4med2} reported that depending on the medical imaging dataset, it can, in some cases, perform accurate medical image segmentation. These experiments also fine tune SAM to the DyABD dataset. The following models were investigated;

\textbf{SAM.} \hspace{0.06cm}  SAM \citep{sam} consists of a Vision Transformer \citep{vit} to encode the image, a prompt encoder and mask decoder. The prompt encoder can take points, boxes, masks, or text as input to guide the segmentation. The mask decoder uses cross-attention to generate a segmentation mask that aligns with the prompts. SAM achieves its generalisation capabilities by training on a large dataset of over 11 million natural images and over 1 billion masks. 

\textbf{SAM 2.} \hspace{0.06cm}  SAM 2 is the second iteration of the SAM model. Although SAM 2 focuses on video segmentation, it has demonstrated improved image segmentation performance and faster inference times by integrating a number of architectural enhancements including replacing the Vision Transformer with a next-generation Hierarchical Vision Transformer \citep{hiera}, allowing the model to identify small and complex objects.

\textbf{MedSAM.} \hspace{0.06cm} MedSAM \citep{medsam} has leveraged SAM's foundational capabilities and has enhanced its segmentation performance by fine tuning it to 1,570,263 medical image-mask pairs, where over 500,000 of those images are MRIs. It was shown to have better accuracy and robustness than modality-wise specialist models.

\textbf{Fine Tuning SAM.} \hspace{0.06cm} This work investigates if fine tuning SAM to the DyABD dataset can improve segmentation performance. During training, each slice of the MRI was input into the image encoder and its bounding box annotation was input into the prompt encoder. The model learns to use the prompt as guidance to segment the abdominal muscles.

For implementation on the DyABD dataset, the coordinates of the bounding boxes were generated based on the segmentation mask annotations. The slices of each dynamic MRI were upsized from $416 \times 416$ to $1024 \times 1024$ and each slice was stacked three times resulting in a final input size of $1024 \times 1024 \times 3$ to align with the input size requirements of the models. For SAM, the largest SAM model, ViT-H was employed, while the SAM 2.1 Hiera Base Plus checkpoint was employed for SAM 2. As fine tuning SAM is computationally intensive, the smaller ViT-B was employed. The hyper-parameters for fine tuning SAM were selected based on the validation set. The model was trained with a learning rate of $1e-6$ and a batch size of two for 65 epochs. Inference for all models was conducted in 2D, meaning that each slice and each muscle was inferred on separately.

\subsection{Fully Automatic} \label{sec:fss_fully_auto}
A fully automated approach to the segmentation of abdominal muscles in dynamic MRI eliminates any annotation cost at test time. However, fully automated approaches are generally trained in a supervised manner which are limited by a requirement for an abundant amount of training data. This analysis investigates the performance of SOTA segmentation models on a novel dataset in the limited data setting.

\textbf{nnU-Net.} \hspace{0.06cm} The nnU-Net \citep{nnunet} is based on the U-Net \citep{unet} encoder-decoder architecture. It achieves SOTA medical image segmentation performance across a wide range of tasks such as tumor and organ segmentation. It is self-configuring algorithm that automatically performs pre-processing (adjusting image resolution, normalization, and intensity scaling), optimisation of the architecture and it also identifies the optimal training configuration such as batch size and learning rate to suit the specific dataset. The 2D implementation treats each slice independently, while the 3D version processes the dynamic MRI volumes as a single input.

\textbf{U-Mamba.} \hspace{0.06cm} U-Mamba \citep{umamba} is also a self-configuring network with a U-Net architecture, similar to the nnU-Net. It has been shown to outperform the nnu-Net in some tasks such as the segmentation of the organs in abdominal MRI and CT. U-Mamba achieves this performance by integrating their CNN-SSM block into each layer of the architecture. The CNN-SSM Block integrates features from CNNs and State Space Models. State Space Models are deep sequence models that are capable of handling long sequences, making them useful for the segmentation of 3D medical imaging data. 

Images were downsized from $416 \times 416$ to $256 \times 256$ for training. The models are self-configured using a validation set. The train to validation set was 80:20. For both nnU-Net and U-Mamba, augmentations relating to flipping, mirroring or rotating were switched off as it was found in previous work \citep{joppin_midl} that such augmentations hindered the model's performance due to the symmetry of the abdominal muscles.

\section{Evaluation}\label{sec:eval}
The two most commonly used metrics in the field of segmentation are the Dice Coefficient and Intersection over Union (IoU), also known as the Jaccard Index. Both metrics are calculated by comparing the segmentation model's output to the ground truth segmentations. The equation for Dice Coefficient is shown in equation \ref{eq:background_dice}, while the equation for IoU is shown in equation \ref{eq:background_iou}, where TP is the number of pixels that were correctly segmented, FP is the number of pixels that were incorrectly segmented and FN is the number of pixels that were incorrectly not segmented. Dice Coefficient and IoU provide measures of the overlap between two segmentations. In contrast, another common segmentation metric, Hausdorff Distance indicates the boundary accuracy of a segmentation. The Symmetric Hausdorff Distance is calculated as in equation \ref{eq:background_haus}, where $GT$ denotes the set of coordinates of the boundary of the ground truth segmentation, $P$ denotes the set of coordinates of the boundary of the segmentation model and $h()$ is the Directed Hausdorff Distance, calculated as in in equation \ref{eq:background_haus2}. To compute the directed Hausdorff distance from $GT$ to $P$, the minimum Euclidean distance is calculated from each point $gt \in GT$ to the closest point in $P$. The largest of these minimum distances is then taken as the directed distance. The Hausdorff distance, therefore, quantifies the maximum boundary deviation between the prediction and ground truth. Higher values indicate poorer segmentation accuracy.


\begin{equation}
\label{eq:background_dice}
    Dice\ Coefficient = \frac{TP \times 2}{(TP+FP) + (TP+FN)}
\end{equation}

\begin{equation}
\label{eq:background_iou}
    IoU = \frac{Dice\ Coefficient}{2-Dice\ Coefficient}
\end{equation}

\begin{equation}
\label{eq:background_haus}
    Symmetric\ Hausdorff\ Distance = max(h(GT,P), h(P,GT))
\end{equation}

\begin{equation}
\label{eq:background_haus2}
h(GT, P) = \max_{gt \in GT} \min_{p \in P} \| gt - p \| 
\end{equation}


\section{Results}\label{sec:res}
This section presents the results of the FSS, Zero Shot and supervised approaches. Each section reports the Dice Coefficient and IoU for each model. The results are broken down by performance on the segmentation of each of the four abdominal muscles, rLM, rRA, lRA and lLM.  When training was required for the model, four fold validation was employed. The results presented are the average performance over the four folds. One standard deviation is reported across folds. For techniques that do not require training such as the Zero Shot methods, they were evaluated in this manner too, meaning the performance was calculated per the same training folds as other techniques and averaged.

\subsection{Semi-Automatic with Annotations}
Table \ref{tab:fss_few_results} presents the Dice Coefficient and IoU for UniverSeg and ADNet, averaged over each fold and each muscle with varying numbers of shots. These techniques were evaluated solely on the slices that were not used as shots in the Support Set. The number of shots were studied up to five as increasing past five results in large portions of the dynamic MRI volumes being annotated. As anticipated, the performance of UniverSeg improves as the number of shots increases, with a particularly notable improvement observed when transitioning from one shot to two shots. In contrast, ADNet does not exhibit a comparable performance gain, suggesting that it is less effective at leveraging information from multiple slices to guide segmentation.

Table \ref{tab:fss_few_broken_results} provides a breakdown of the results, along with the Hausdorff Distance for models that trained with five shots by each muscle. Despite requiring no training, UniverSeg outperforms ADNet by substantial margins, highlighting its generalisation capabilities. UniverSeg demonstrates consistent performance across all muscles, while ADNet has reduced performance on the RA muscles, in terms of Dice Coefficient and IoU. The RA muscles are more challenging to segment due to the greater range of motion. Figure \ref{fig:fss_seg_res2} shows the qualitative examples of the model output. ADNet struggles to delineate precise boundaries, while UniverSeg produces highly accurately segmentations for coughing and Valsalva, however it makes substantial errors when segmenting MRIs performing the breathing exercise.
 

\begin{table}
  \caption[Dice Coefficient and IoU for Varying Numbers of Shots Averaged over each Fold for Semi-Automatic with Annotation Methods.]{Dice Coefficient and IoU for varying numbers of shots averaged over each fold for semi-automatic with annotation methods. }
  \label{tab:fss_few_results}

  \centering
  \begin{tabular}{l|ccccc}
    \hline
\hline
  Method &  \multicolumn{1}{|c}{1 Shot}  &  \multicolumn{1}{c}{2 Shots}  
  &  \multicolumn{1}{c}{3 Shots}  &  \multicolumn{1}{c}{4 Shots} &  \multicolumn{1}{c}{5 Shots} 
      \\
      \hline
      &\multicolumn{4}{c}{\textbf{Dice}}\\

       ADNet  &  $\mathbf{0.61 \pm{} 0.07}$& $0.66\pm{}0.03$ & $0.66\pm{}0.04$ & $0.65\pm{} 0.04$  & $0.64\pm{} 0.04$  \\
   
     UniverSeg &  $0.41 \pm{} 0.01$& $\mathbf{0.78\pm{}0.01}$ & $\mathbf{0.81\pm{}0.02}$ & $\mathbf{0.81\pm{} 0.02}$ & $\mathbf{0.82\pm{} 0.02}$  \\

     \hline 
           &\multicolumn{4}{c}{\textbf{IoU}}\\

               ADNet &  $\mathbf{0.50 \pm{} 0.06}$& $0.51\pm{}0.03$ & $0.51\pm{}0.05$ & $0.50\pm{} 0.05$ & $0.49\pm{} 0.05$  \\
                 UniverSeg  &  $0.28 \pm{} 0.01$& $\mathbf{0.66\pm{}0.01}$ & $\mathbf{0.69\pm{}0.02}$ & $\mathbf{0.70\pm{} 0.02}$  & $\mathbf{0.70\pm{} 0.02}$\\

  \end{tabular}
  
  \end{table}

\begin{table}
  \caption[Dice Coefficient, IoU and Hausdorff Distance Averaged over each Fold for Semi-Automatic with Annotation Methods.]{Dice Coefficient, IoU and Hausdorff Distance averaged over each fold for Semi-automatic with annotation methods trained with five shots. }
  \label{tab:fss_few_broken_results}

  \centering
  \begin{tabular}{l|ccccc}
    \hline
\hline
  Method &  \multicolumn{1}{|c}{rLM}  &  \multicolumn{1}{c}{rRA}  
  &  \multicolumn{1}{c}{lRA}  &  \multicolumn{1}{c}{lLM}  &  \multicolumn{1}{c}{Average}  
      \\
      \hline
      &\multicolumn{5}{c}{\textbf{Dice}}\\

     ADNet &  $0.72 \pm{}0.06$& $0.57\pm{}0.07$ & $0.58\pm{}0.04$ & $0.74\pm{}0.06 $ & $0.65\pm{}0.05$ \\

     UniverSeg  &  $\mathbf{0.79 \pm{} 0.07}$& $\mathbf{0.82\pm{}0.01}$ & $\mathbf{0.84\pm{}0.01}$ & $\mathbf{0.82\pm{} 0.03}$ & $\mathbf{0.82\pm{}0.02}$ \\

     \hline 
           &\multicolumn{5}{c}{\textbf{IoU}}\\
    ADNet &  $0.58 \pm{}0.06$& $0.41\pm{}0.07$ & $0.41\pm{}0.04$ & $0.60\pm{}0.07 $ & $0.50\pm{}0.06$ \\

      UniverSeg  & $\mathbf{0.68\pm{}0.08}$ & $\mathbf{0.70\pm{}0.02}$ & $\mathbf{0.73\pm{}0.02}$  & $\mathbf{0.71\pm{}0.03}$ & $\mathbf{0.70\pm{}0.02}$ \\

 \hline 
           &\multicolumn{5}{c}{\textbf{Hausdorff Distance}}\\

    ADNet &  $35.75 \pm{}4.47$& $14.26\pm{}3.45$ & $13.64\pm{}1.96$ & $34.77\pm{}4.25 $ & $24.60\pm{}1.93$ \\

      UniverSeg  & $\mathbf{25.58\pm{}10.93}$ & $\mathbf{8.93\pm{}3.62}$ & $\mathbf{9.92\pm{}4.11}$  & $\mathbf{19.22\pm{}4.46}$ & $\mathbf{15.91\pm{}5.29}$ \\
       
  \end{tabular}
  
  \end{table}

\subsection{Semi-Automatic with Prompts}

\begin{figure}[H]
\centerline{\includegraphics[width=\textwidth]{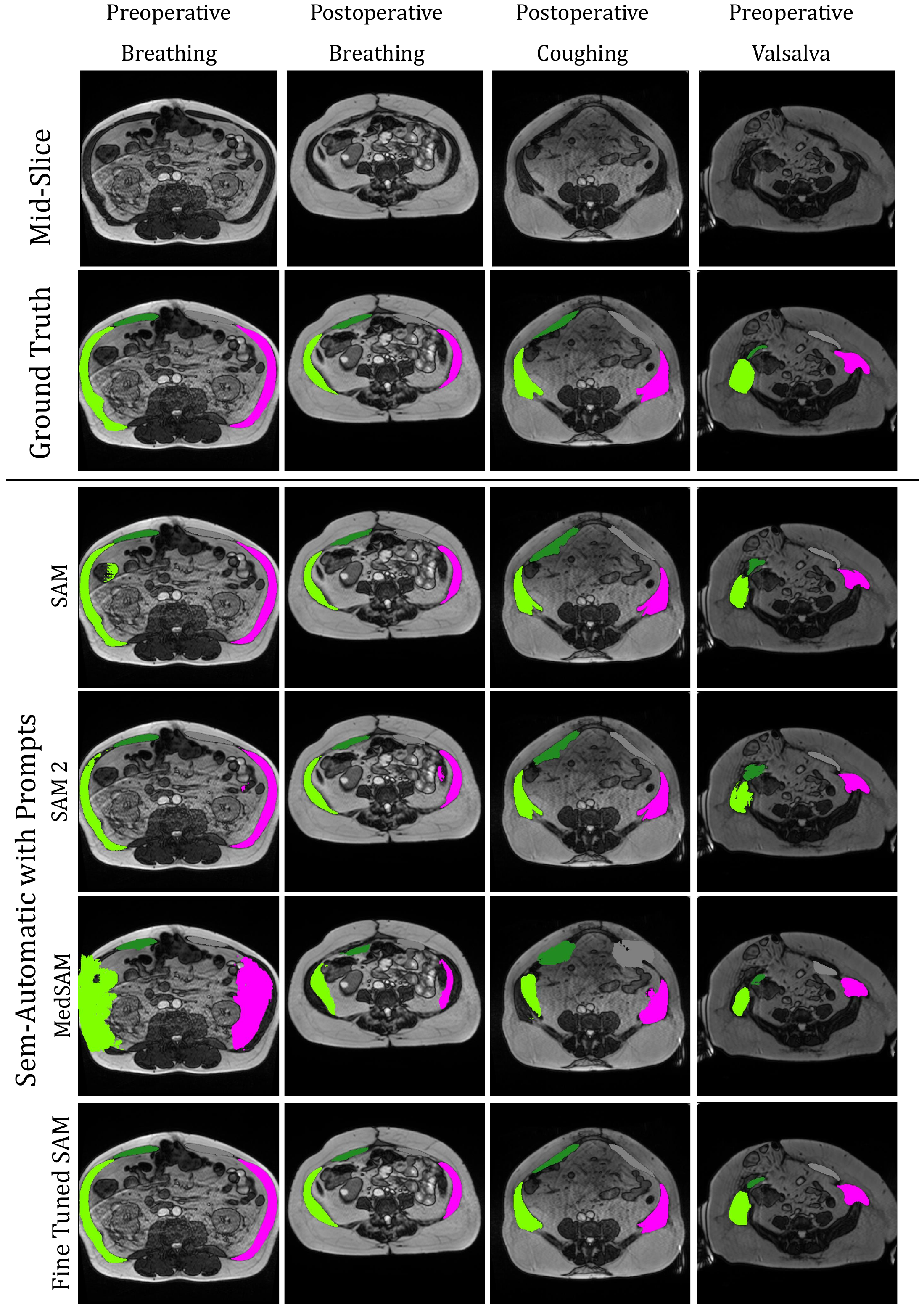}}
\caption[Visual Segmentation Results for the Approaches; Semi-Automatic with Annotations, Fully Automatic and Semi-Automatic with Prompts.]{Visual segmentation results for semi-automatic with annotations, fully automatic and semi-automatic with prompts. Models are shown row-wise and the mid-slice from four randomly sampled patients are shown column-wise.}
\label{fig:fss_seg_res2}
\vspace{0.06cm}
\end{figure}

Table \ref{tab:fss_zero_results} presents the Dice Coefficient, IoU and Hausdorff Distance for the semi-automatic methods with prompting, averaged over each fold. SAM, SAM2, and MedSAM do not require training but rely on a bounding box prompt during inference. Fine-tuned SAM, on the other hand, requires both training and bounding box prompts at inference. Despite MedSAM being training specifically for medical imaging data, the foundation models SAM and SAM2 demonstrate superior performance for abdominal muscle segmentation, suggesting that MedSAM is less generalisable to unseen modalities and tasks. The qualitative examples of MedSAM in figure \ref{fig:fss_seg_res2}, demonstrates that it can fail to identify the abdominal muscle in some cases (lLM muscle in postoperative breathing). SAM and SAM2 achieve comparable results in terms of Dice Coefficient and IoU. However, it is evident from the Hausdorff Distance that SAM demonstrates higher boundary accuracy. Fine-tuning SAM leads to large performance improvements, achieving an average Dice Coefficient value of 0.89 but at an increased annotation cost. The qualitative examples in figure \ref{fig:fss_seg_res2} show almost perfect performance for fine tuned SAM. Additionally, results are consistently higher for LM muscles across all SAM models, which is the anticipated result as there is more muscle movement in the RA muscles.

\begin{table}
  \caption[Dice Coefficient, IoU \textcolor{red}{and Hausdorff Distance} Averaged over each Fold for Semi-Automatic Methods with Prompts]{Dice Coefficient, IoU and Hausdorff Distance averaged over each fold for semi-automatic methods with prompts.}
  \label{tab:fss_zero_results}

  \centering
  \begin{tabular}{l|ccccc}
    \hline
\hline
  Method &  \multicolumn{1}{|c}{rLM}  &  \multicolumn{1}{c}{rRA}  
  &  \multicolumn{1}{c}{lRA}  &  \multicolumn{1}{c}{lLM}  &  \multicolumn{1}{c}{Average}  
      \\
      \hline
      &\multicolumn{5}{c}{\textbf{Dice}}\\

     SAM  &  $0.88 \pm{} 0.03$  & $0.77\pm{} 0.06$  &  $0.79\pm{} 0.09$  & $0.87\pm{} 0.02$ &  $0.83\pm{} 0.04$  \\

     SAM2  &   $0.85 \pm{} 0.05$ & $0.78\pm{}0.08$ & $0.80\pm{}0.09$ & $0.85 \pm{}0.05$ & $0.82\pm{}0.05$ \\

     MedSAM  &  $0.57\pm{} 0.06$  & $0.52\pm{} 0.10$ & $0.57\pm{} 0.08$ &$0.53 \pm{} 0.07$ & $0.55\pm{} 0.07$ \\

     Fine Tuned SAM &  $\mathbf{0.91 \pm{} 0.01}$  &  $\mathbf{0.86\pm{}0.02}$ & $\mathbf{0.88\pm{}0.02}$ & $\mathbf{0.92\pm{} 0.01}$ & $\mathbf{0.89\pm{}0.01}$ \\
     
     \hline 
           &\multicolumn{5}{c}{\textbf{IoU}}\\
    SAM &  $0.79 \pm{} 0.04$  & $0.65\pm{} 0.08$  & $0.68\pm{} 0.10$ & $0.78\pm{} 0.03$ & $0.72\pm{} 0.04$  \\

     SAM2  &$0.75 \pm{} 0.07$  & $0.66\pm{} 0.09$ & $0.70\pm{}0.10$& $0.75\pm{}0.07$ & $0.71\pm{}0.06$ \\

     MedSAM &$0.43 \pm{}0.06$  & $0.39\pm{}0.10$  & $0.44\pm{}0.07$  & $0.40\pm{}0.07$ & $0.41\pm{}0.07$ \\
      Fine Tuned SAM  &$\mathbf{0.84\pm{} 0.02}$ & $\mathbf{0.76\pm{}0.03}$ & $\mathbf{0.78\pm{}0.03}$ & $\mathbf{0.85\pm{}0.02}$ & $\mathbf{0.81\pm{}0.02}$\\
            \hline 
           &\multicolumn{5}{c}{\textbf{Hausdorff Distance}}\\
    SAM &  $14.97 \pm{} 4.53$  & $11.44\pm{} 3.77$  & $9.33\pm{} 5.01$ & $16.20\pm{} 4.84$ & $12.99\pm{} 3.33$  \\

     SAM2  &$19.87 \pm{} 6.80$  & $22.68\pm{} 13.77$ & $16.13\pm{}10.96$& $25.10\pm{}9.28$ & $20.95\pm{}7.06$ \\

     MedSAM &$27.06 \pm{}3.22$  & $17.86\pm{}2.39$  & $17.87\pm{}3.02$  & $31.02\pm{}5.46$ & $23.45\pm{}2.99$ \\
    Fine Tuned SAM  &$\mathbf{11.34\pm{} 1.73}$ & $\mathbf{9.98\pm{}1.16}$ & $\mathbf{9.31\pm{}1.28}$ & $\mathbf{12.08\pm{}1.69}$ & $\mathbf{10.68\pm{}0.99}$\\
  \end{tabular}
  
  \end{table}

\subsection{Fully Automatic}
Table \ref{tab:fss_supp_results} presents the results of the fully automatic supervised approaches, nnU-Net (2D), nnU-Net (3D) and U-Mamba. Despite supervised methods being designed for settings where there is an abundant amount of training data and annotations available, all models generalise well to the DyABD dataset. The nnU-Net (3D) achieves the highest segmentation performance, achieving an average Dice Coefficient of 0.82. The superior performance is likely due to the 3D architecture which can capture inter-slice relationships.  Notably, this result is achieved despite the increased architectural complexity of 3D models, which often makes them more susceptible to overfitting. These findings suggest that even with a relatively small dataset, 3D architectures represent a promising direction for abdominal muscle segmentation in dynamic MRI. Figure \ref{fig:fss_seg_res2} presents qualitative examples of the model's segmentations.

\begin{table}
  \caption[Dice Coefficient, IoU and Hausdorff Distance Averaged over each Fold for Fully Automated Datasets.]{Dice Coefficient, IoU and Hausdorff Distance averaged over each fold for fully automated methods. }
  \vspace{0.06cm}
  \label{tab:fss_supp_results}

  \centering
  \begin{tabular}{l|ccccc}
    \hline
\hline
  Method &  \multicolumn{1}{|c}{rLM}  &  \multicolumn{1}{c}{rRA}  
  &  \multicolumn{1}{c}{lRA}  &  \multicolumn{1}{c}{lLM}  &  \multicolumn{1}{c}{Average}  
      \\
      \hline
      &\multicolumn{5}{c}{\textbf{Dice}}\\

          nnU-Net (2D) &   $0.84\pm{} 0.07$ &  $0.72\pm{}0.09$ & $0.72\pm{}0.11$ & $0.83\pm{} 0.07$ &  $0.78\pm{} 0.08$  \\
        nnU-Net (3D) &  $\mathbf{0.87 \pm{} 0.04}$ & $\mathbf{0.77\pm{} 0.06}$& $\mathbf{0.76\pm{} 0.11}$ &  $\mathbf{0.89\pm{} 0.02}$  & $\mathbf{0.82\pm{} 0.05}$ \\
              U-Mamba &  $0.82 \pm{}0.05$  & $0.71\pm{}0.07$& $0.73\pm{}0.11$ &  $0.84 \pm{}0.05$  &  $0.78 \pm{}0.06$  \\

     \hline 
           &\multicolumn{5}{c}{\textbf{IoU}}\\
          nnU-Net (2D)  &  $0.75 \pm{}0.06$  &  $0.59 \pm{}0.10$  &  $0.60\pm{}0.12$ & $0.78\pm{}0.07$ &  $0.67\pm{}0.08$ \\
        nnU-Net (3D) &  $\mathbf{0.79 \pm{}0.04}$ &  $\mathbf{0.65\pm{} 0.07}$  &  $\mathbf{0.65\pm{}0.12}$ & $\mathbf{0.81\pm{} 0.03} $ &  $\mathbf{0.72\pm{}0.06}$ \\
            U-Mamba &   $0.73 \pm{}0.05$ &  $0.58 \pm{}0.09$ &  $0.61\pm{}0.12$ & $0.75\pm{}0.06$ &  $0.67\pm{}0.07$ \\
         \hline 
           &\multicolumn{5}{c}{\textbf{Hausdorff Distance}}\\

        nnU-Net (2D)  &  $47.46 \pm{}20.33$  &  $33.95 \pm{}13.69$  &  $28.59\pm{}19.04$ & $42.37\pm{}25.34$ &  $38.09\pm{}17.55$ \\
    nnU-Net (3D) &  $\mathbf{16.76 \pm{}4.48}$ &  $\mathbf{15.34\pm{} 4.12}$  &  $\mathbf{16.88\pm{}8.97}$ & $\mathbf{16.00\pm{} 5.11} $ &  $\mathbf{16.24\pm{}4.96}$ \\
        U-Mamba &   $39.01 \pm{}18.89$ &  $30.35 \pm{}21.19$ &  $25.35\pm{}13.92$ & $32.04\pm{}12.24$ &  $31.69\pm{}16.18$ \\

  \end{tabular}
  
  \end{table}

\subsection{Comparison of Approaches}
Table \ref{tab:fss_sup_results} summarises the Dice Coefficient of all approaches to the segmentation of abdominal muscles in dynamic MRI. The table also specifies the level of automation and the training requirements of each technique. When analysing the results, it should be noted that the direct comparison of FSS models UnivSeg and ADNet to other approaches is limited by the fact that they are only evaluated on the portion of slices that were not present in the Support Set.

This analysis found that the fully supervised nnU-Net (3D) generalises well to the DyABD dataset, achieving a Dice Coefficient of 0.82. This is a notable result as it was achieved having trained on a small sample of patients and supervised approaches are designed for the setting where there is abundant training data available. They are generally considered to be at risk of overfitting when trained on a small dataset. Furthermore, it substantially outperforms ADNet, which is trained on the same portion of data as the nnU-Net but also receives additional supervision at inference time in the form of a Support Set. This finding underscores that, fully supervised models can exceed the performance of FSS models, even in data-limited settings.

The supervised approach of nnU-Net however performs on par with the promptable Zero Shot techniques, SAM and SAM 2. This suggests a trade-off between the two methods; the supervised nnU-Net requires training data with precise annotations of the complete dataset as a one-time effort, while SAM and SAM2 are consistently reliant on coarser bounding box annotations at inference.

UniverSeg also performs at the same level as nnU-Net (3D) and the promptable Zero Shot methods, achieving a dice score of 0.82. There is also an additional trade-off between the FSS method UniverSeg and the promptable Zero Shot SAM models. UniverSeg requires the precise delineation of the abdominal muscles on a small portion of the data, while the promptable Zero Shot methods require coarse annotations of the complete dataset. An advantage of the latter is that bounding box annotations could be performed by a non expert with brief training, while the former would require an expert, thus indicating that the promptable Zero Shot methods have the lower annotation cost.

The best performing method by a substantial margin was the fine-tuned version of SAM. However, this approach comes with the highest annotation cost as it requires precise annotations for training, and bounding box prompts for both training and inference.

A paired t-test was performed to determine if the difference in results between each pair of models was statistically significant. This was conducted on the best performing models from each class, UniverSeg (semi-automatic with annotations), nnU-Net (3D) (fully automatic), and SAM and fine tuned SAM (semi-automatic with prompts). It was found that the performance of all models have a statistically significant difference with the exception of SAM and UniverSeg $(p=0.481)$, and additionally SAM and nnU-Net $(p=0.608)$. This is anticipated as these pairs of models achieve similar Dice Coefficients. 


To conclude, FSS model UniverSeg, promptable Zero Shot methods SAM and SAM 2, and supervised model nnU-Net (3D) have demonstrated that they can generalise to the task of abdominal muscle segmentation They on average achieve a Dice Coefficient of 0.82, meaning there is still potential to further improve the segmentation of these techniques.

\begin{table}
  \caption[Summary of Results.]{Summary of results (Dice Coefficient) for each approach including details on their level of automation and training requirement.}
  \label{tab:fss_sup_results}

  \centering
  \begin{tabular}{l|ccc}
    \hline
\hline
     \multicolumn{1}{c|}{Technique}  &  \multicolumn{1}{c}{Category} &  \multicolumn{1}{c}{Requires Training}  &   \multicolumn{1}{c}{Performance} 
      \\
      \hline
       
          UniverSeg  &Semi-Automatic with Annotations & No &  0.82\\
        ADNet     & Semi-Automatic with Annotations & Yes & 0.64  \\
        
         SAM   & Semi-Automatic with Prompts &No     & 0.83\\

          SAM2  & Semi-Automatic with Prompts & No   & 0.82 \\
        
           MedSAM   & Semi-Automatic with Prompts &No  &   0.55 \\

          Fine Tuned SAM  &Semi-Automatic with Prompts &  Yes    & \bfseries 0.89  \\

          nnU-Net (2D)    &Fully-Automatic &  Yes   & 0.78\\

       nnU-Net (3D)    &  Fully-Automatic & Yes     & 0.82\\
        
       U-Mamba  & Fully-Automatic &  Yes    & 0.78\\

  \end{tabular}
  
  \end{table}

\subsection{Results by Exercise and Operative Stage}
\begin{figure*}[!t]
\centerline{\includegraphics[width=\textwidth]{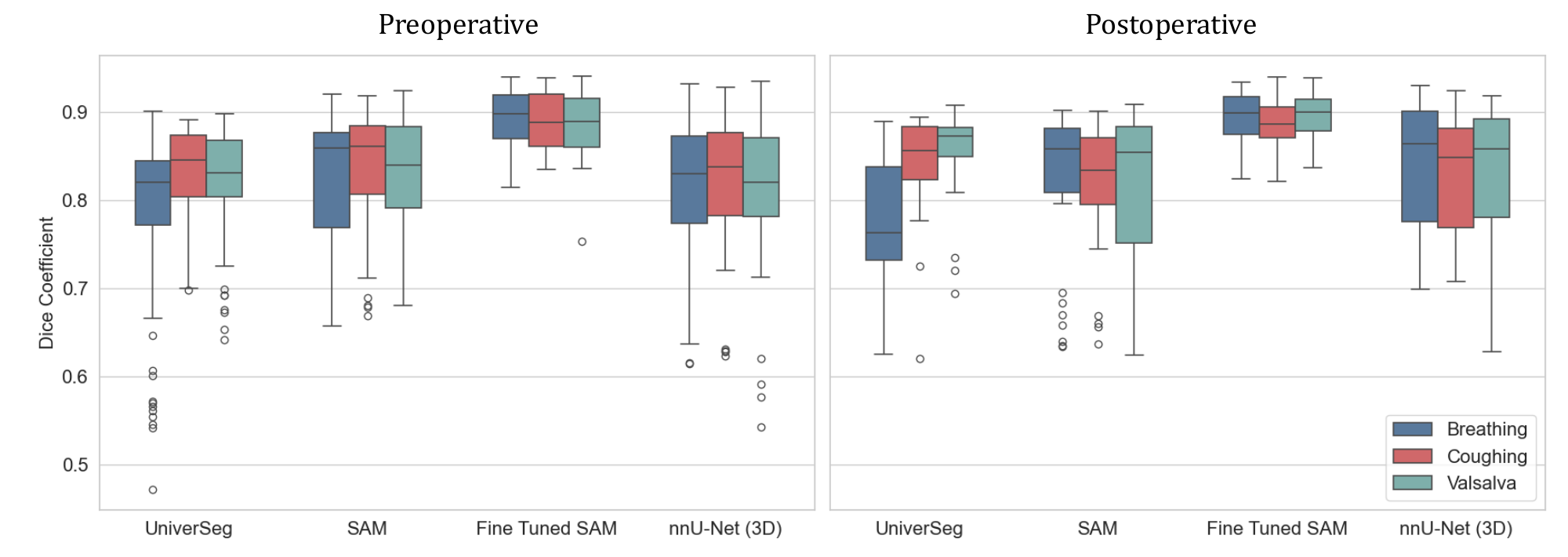}}
\caption[Dice Coefficient By Exercise and Operative Stage for Various Models.]{Dice Coefficient by exercise and operative stage for UniverSeg (semi automatic with annotations), nnU-Net (3D) (fully automatic), SAM and fine tuned SAM (semi automatic with prompts).}
\label{fig:fss_ex_op}
\vspace{0.06cm}
\end{figure*}

This section analyses the results by exercise and operative stage for each of the best performing models within each category, UniverSeg (semi automatic with annotations), nnU-Net (3D) (fully automatic), SAM and fine tuned SAM (semi automatic with prompts). Figure \ref{fig:fss_ex_op} shows the distribution of Dice Coefficients of the dynamic MRI volumes for each exercise, operative stage and model. 

Statistical tests were conducted to test if there is a difference in the segmentation performance between preoperative and postoperative stages. A paired t-test was conducted between all volumes which had both a preoperative and postoperative MRI (paired data) when the data had a normal distribution and Wilcoxon Signed-Rank Test otherwise. It was found that there was a statistically significant difference between the performance of UniverSeg for cases performing the Valsalva maneuver preoperatively versus postoperatively $(p=0.010)$. It can be seen from the figure that UniverSeg performs better on the Valsalva maneuver post corrective surgery. However, there was no statistically significant difference between preoperative and postoperative stages for any other model, indicating that generally segmentation models perform uniformly on MRIs acquired preoperative and postoperative.

An additional statistical test was conducted to test if there is a difference in the segmentation performance for different exercises. For each model and each operative stage, a paired t-test was performed when the data had a normal distribution and Wilcoxon Signed-Rank Test otherwise. It was found that there was a statistically significant difference between the performance of UniverSeg for cases performing the breathing exercise and cases performing the Valsalva maneuver $(p=0.013)$. The differences in the Dice Coefficient distributions can evidently be seen in the figure. The qualitative examples in figure \ref{fig:fss_seg_res2} highlight UniverSeg's poorer performance on the breathing exercise. This is in contrast to the expected result as less muscle movement occur during the breathing exercise.

\subsection{Patient Variability}
The segmentation results previously shown were averaged across folds. The aim of this section is to more closely analyse the variability of results between patients. Table \ref{tab:fss_pat} reports the average Dice Coefficient averaged for each patient across exercises, operative stages and folds for each of the best performing models of each category, UniverSeg (semi automatic with annotations), nnU-Net (3D) (fully automatic), SAM and fine tuned SAM (semi automatic with prompts). One standard deviation calculated across patients is also shown. Based on the lower standard deviation, it can be observed that UniverSeg and fine tuned SAM have less inter-patient variability, likely due to the prompt guidance at inference, while there is higher inter-patient variability for nnU-Net which receives no level of supervision at inference.

\begin{table}
  \caption[Results Averaged Across Patients.]{Dice Coefficient with one standard deviation (std), averaged across patients.}
  \label{tab:fss_pat}

  \centering
  \begin{tabular}{l|c}
    \hline
\hline
     \multicolumn{1}{c|}{Technique}  &  \multicolumn{1}{c}{Dice Coefficient $\pm{}std$}
      \\
      \hline
       
          UniverSeg  &0.82 $\pm{}0.04$\\
        
         SAM   & 0.83$\pm{}0.06$\\

          Fine Tuned SAM & \textbf{0.89 $\pm{}0.03$} \\

       nnU-Net (3D)    &   0.82$\pm{}0.07$\\

  \end{tabular}
  
  \end{table}

\section{Discussion and Conclusion}\label{sec:fss_concl}
This work introduced the novel, first-of-its-kind abdominal muscle segmentation benchmark dataset, DyABD. DyABD is unlike any other existing benchmark dataset in that it contains MRIs of preoperative and postoperative stages and it was acquired dynamically while patients performed various different exercises. Given the success of various organised challenges such as The Liver Tumor Segmentation Benchmark (LiTS) \citep{lits}, The Multimodal Brain Tumor Segmentation Challenge \citep{brats3} and The Medical Segmentation Decathlon \citep{med_decath} at the Medical Image Computing and Computer Assisted Intervention Society (MICCAI) conference, future work will focus on organising a public abdominal muscle segmentation challenge to further promote research in this area.

This work also shed light on the true progress of the field in medical image segmentation by comprehensively evaluating the generalisation capabilities of existing methods to the complex and challenging dataset, DyABD. This analysis found that FSS, promptable Zero Shot and fully supervised methods achieve Dice Coefficients $\ge$ 0.82 on the task of abdominal muscle segmentation in dynamic MRI. However, there is still room for substantial improvement. The 3D supervised approach, nnU-Net performs comparably with FSS and promptable Zero Shot models, despite being originally designed for settings with abundant training data. Moreover, the supervised model receives no level of supervision at inference, while FSS and promptable Zero Shot models are guided by a Support Set or a prompt. However, nnU-Net has a training requirement, while FSS model, UniverSeg and promptable Zero Shot models SAM and SAM2 can be used off-the-shelf, resulting in a trade-off between a once-off high annotation cost for training the nnU-Net or a consistent but smaller annotation cost at inference time for FSS and promptable Zero Shot models. The best performing model in the limited data setting was fine tuning the promptable Zero Shot model, SAM, which increased the segmentation performance by substantial margins, achieving a Dice Coefficient of 0.89. However this comes at an increased annotation cost as it requires annotated training data and prompts at inference time. 

This analysis also found that there was no evident difference in the segmentation performance between preoperative and postoperative dynamic MRIs for the majority of models. The same result was found when comparing the performance between exercises, with the exception of UniverSeg which had poorer performance on the breathing exercise. An additional finding was that the models had poorer performance on the segmentation of RA muscles in comparison to the LM muscles. This likely due to the increased movement of the RA muscles (figure \ref{fig:fss_radial_dist2}).

While the studied methods demonstrated the generalisation capabilities on the task of abdominal muscle segmentation, there is still potential for further performance improvements. Given the observed performance improvement of 3D nnU-Net over 2D nnU-Net, future work could include the further exploration of 3D and spatio-temporal video segmentation techniques that can model the relationships between slices. There have been various techniques proposed that are suitable for 3D medical image segmentation such as transformers \citep{cotr,transmed}, volumetric attention mechanisms \citep{volatt}, along with other Mamba based models \citep{segmamba}, all of which are capable of modelling sequential data. Additionally, \cite{medvid} demonstrated that the video implementation of SAM 2 \citep{sam2} can perform accurate medical image segmentation on a range of 2D and 3D tasks including abdominal organ segmentation. Another promising future direction is developing bounding box prompt propagation techniques to minimise the annotation cost associated with promptable Zero Shot models. At present, each slice of each MRI requires a bounding box annotation, however bounding box coordinates from one slice could be sufficient for multiple consecutive slices or require only minor modification. While some work has investigated automated approaches to prompting \citep{autosam}, they focus on key point detection, rather than bounding box propagation.

Future work will be conducted on the downstream task of automating the quantification of the abdominal wall motion by calculating key metrics such as radial displacement \citep{dynamic_mri,vic}. This shifts the focus of the proposed study from optimising the Dice Coefficient to evaluating the model's segmentation performance based on which method derives the most accurate abdominal wall quantification metrics, thus ensuring there is clinical relevance to the segmentation task. Furthermore, previous preliminary work  on generating synthetic abdominal MRIs showed promising results \citep{dyabd}. Therefore, future work will also focus on artificially generating adbominal dynmaic MRIs with the goal of overcoming various challenges such as the small-medium dataset size.

While this work focuses on the limited data setting, one limitation lies in the testing phase, as drawing definitive conclusions based on a small test set can be challenging. However, this was mitigated in the experimental setup by employing four-fold cross-validation, which ensured that the entire dataset was utilised for testing. Another limitation is the presence of Motion Artefacts in a portion of the dynamic MRIs, caused by muscle movement during acquisition. These artefacts can obscure the precise boundaries of the muscles, making annotation and model segmentation more challenging. 

\begin{figure}[H]
\centerline{\includegraphics[width=\textwidth]{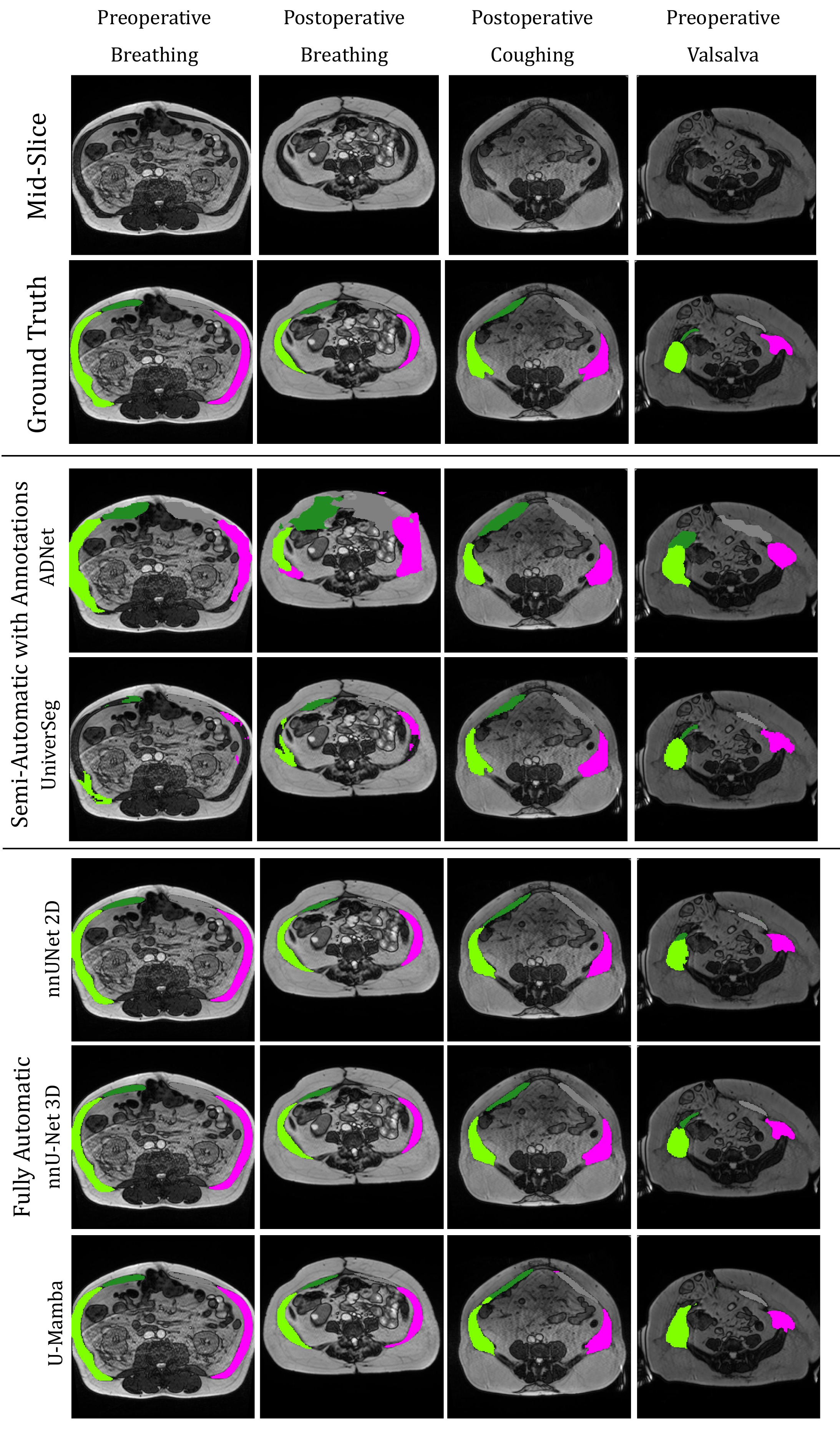}}

\label{fig:fss_seg_res}
\vspace{0.06cm}
\end{figure}

\section*{Declarations}
\subsection*{Abbreviations}

\begin{description}[leftmargin=!,labelwidth=2cm]
    \item[\textbf{CNN}] Convolutional Neural Network
    \item[\textbf{CT}] Computed Tomography
    \item[\textbf{DL}] Deep Learning
    \item[\textbf{FSS}] Few Shot Segmentation
    \item[\textbf{IoU}] Intersection over Union
    \item[\textbf{LM}] Lateral Muscle
    \item[\textbf{lLM}] left Lateral Muscle
    \item[\textbf{lRA}] left Rectus Abdominus
    \item[\textbf{MRI}] Magnetic Resonance Imaging
    \item[\textbf{RA}] Rectus Abdominus
    \item[\textbf{rLA}] right Lateral Muscle
    \item[\textbf{rRA}] right Rectus Abdominus
    \item[\textbf{SAM}] Segment Anything Model
\end{description}

\subsection*{Ethics Approval and Consent to Participate}
This study was approved by the French ethics committee (IDRCB: 2021-A02119-32) and was conducted at Assistance Publique des Hôpitaux de Marseille (APHM), France according to national legislation related to interventional research and the Declaration of Helsinki. Informed consent to participate was obtained from all of the participants in the study. The data is anonymised.

\subsection*{Consent for Publication}
Consent for publication of images may is not required as the images are unidentifiable and the dataset is anonymised.

\subsection*{Availability of Data and Materials}
The DyABD dataset is available for request at \url{https://entrepot.recherche.data.gouv.fr/dataset.xhtml?persistentId=doi:10.57745/KTM2OA}. The code for this study will be available at \url{https://github.com/niamhbelton/DyABD-Segmentation}.

\subsection*{Competing Interests}
There are no competing interests to declare.

\subsection*{Funding}
This work was funded by Science Foundation Ireland through the SFI Centre for Research Training in Machine
Learning (Grant No. 18/CRT/6183). This work is supported by the Insight Centre for Data Analytics under Grant Number SFI/12/RC/2289 P2. 

\subsection*{Author contributions}
N.B. performed the data annotation, all experiments and wrote the manuscript.
V.J. conducted the data acquisition, data preprocessing, data annotation, results interpretation and paper editing. 
A.L. provided technical guidance and supervision throughout the experiments. 
C.M. took part in initial study design, data acquisition and advised on the paper.
T.B. took part in initial study design, data acquisition, advised on annotations and advised on the paper.
D.B.   took part in initial study design, data acquisition, results interpretation and advised on the paper. 
K.M.C. took part in study design, results interpretation and advised on the paper. 

\subsection*{Acknowledgements}
This work is supported by the University College Dublin (UCD) AI Healthcare Hub, UCD Institute for Discovery.

\bibliography{sn-bibliography}

\end{document}